\documentclass[11pt,a4paper]{article}
\usepackage[hyperref]{emnlp2018}
\usepackage{times}
\usepackage{latexsym}
\usepackage{graphicx}  
\usepackage{amsmath}
\usepackage{amsfonts}
\usepackage{url}
\usepackage{placeins}
\usepackage{array}
\usepackage{multirow}
\usepackage{booktabs}
\usepackage{enumerate}   
\newcolumntype{?}{!{\vrule width 1.5pt}}
\aclfinalcopy 

\newcommand{\indentitem}{\setlength\itemindent{25pt}}

\title{NEXUS Network: Connecting the Preceding and the Following in Dialogue Generation}

\author{Hui Su$^{1}$\thanks{Indicates equal contribution. X. Shen focuses on algorithm and H. Su is responsible for experiments.}, 
Xiaoyu Shen$^{2,3}\footnotemark[1]
$, Wenjie Li$^4$ and Dietrich Klakow$^3$\\
$^1$Pattern Recognition Center, Wechat, Tencent, China\\
$^2$Max Planck Institute for Informatics, Saarland Informatics Campus, Germany\\
$^3$Spoken Language Systems (LSV), Saarland University, Germany\\
$^4$The Hong Kong Polytechnic University, Hong Kong\\
\tt{aaronsu@tencent.com, xshen@mpi-inf.mpg.de }}

\date{21/08/2018}

\begin{document}
\maketitle
\begin{abstract}
  Sequence-to-Sequence (seq2seq) models have become overwhelmingly popular in building end-to-end trainable dialogue systems. Though highly efficient in learning the backbone of human-computer communications, they suffer from the problem of strongly favoring short generic responses. In this paper, we argue that a good response should smoothly connect both the preceding dialogue history and the following conversations. We strengthen this connection through mutual information maximization. To sidestep the non-differentiability of discrete natural language tokens, we introduce an auxiliary continuous code space and map such code space to a learnable prior distribution for generation purpose. Experiments on two dialogue datasets validate the effectiveness of our model, where the generated responses are closely related to the dialogue context and lead to more interactive conversations.
\end{abstract}

\section{Introduction}
With the availability of massive online conversational data, there has been a surge of interest in building open-domain chatbots with data-driven approaches. Recently, the neural network based sequence-to-sequence (seq2seq) framework~\cite{sutskever2014sequence,cho2014learning} has been widely adopted. In such a model, the encoder, which is typically a recurrent neural network (RNN), maps the source tokens into a fixed-sized continuous vector, based on which the decoder estimates the probabilities on the target side word by word. The whole model can be efficiently trained by maximum likelihood (MLE) and has demonstrated state-of-the-art performance in various domains.
\begin{figure}
\centering
\fbox{\begin{minipage}{0.47\textwidth}
\textcolor{blue}{\textbf{$A_1$}}: Do you know the movie Star Wars?\\
\textcolor{red}{\textbf{$B_1$}}:  Only a bit. \textcolor{purple}{\underline{You can tell me about it!}}\\
\textcolor{blue}{\textbf{$A_2$}}: Of course! This is about ...
\end{minipage}
}
\caption{A conversation in real life}
\label{fig:example}
\end{figure}
However, this architecture is not suitable for modeling dialogues. Recent research has found that while the seq2seq model generates syntactically well-formed responses, they are prone to being off-context, short, and generic (e.g., ``I don’t know" or ``I am not sure")~\cite{li2015diversity,serban2015building}. The reason lies in the one-to-many alignments in human conversations, where one dialogue context is open to multiple potential responses. When optimizing with the MLE objective, the model tends to have a strong bias towards safe responses as they can be literally paired with arbitrary dialogue context without semantical or grammatical contradictions. These safe responses break the dialogue flow without bringing any useful information and people will easily lose interest in continuing the conversation.

In this paper, we propose NEXUS Network which aims at producing more on-topic responses to maintain an interactive conversation flow. Our assumption is that a good response should serve as a ``nexus": connecting and being informative to both the preceding dialogue context and the follow-up conversations. For example, in Figure \ref{fig:example}, the response from $B_1$ is a smooth connection, where the first half indicates the preceding context is a ``Do you know" question and the second half informs that the follow-up would be an introduction about \textit{Star Wars}. We establish this connection by maximizing the mutual information (MMI) of the current utterance with both the past and future contexts. In this way, generic responses can be largely discouraged as they contain no valuable information and thus have only weak correlations with the surrounding context. To enable efficient training, two challenges exist.

The first challenge comes from the discrete nature of language tokens, hindering efficient gradient descent. One strategy is to estimate the gradient by methods like Gumbel-Softmax~\cite{maddison2016concrete,jang2016categorical} or REINFORCE algorithm~\cite{williams1992simple}, which has been applied in many NLP tasks~\cite{he2016dual,shetty2017speaking,gu2017neural,paulus2017deep}, but the trade-off between bias and variance of the estimated gradient is hard to reconcile. The resulting model usually strongly relies on sensitive hyper-parameter tuning, careful pre-train and task-specific tricks. \citet{li2015diversity,wang2017steering} avoid this non-differentiability problem by learning a separate backward model to rerank candidate responses in the testing phase while still adhering to the MLE objective for training. However, the candidate set normally suffers from low diversity and a huge sample size is needed for good performance~\cite{li2016simple}.

The second challenge relates to the unknown future context in the testing phase. In our framework, both the history and future context need to be explicitly observed in order to compute the mutual information. When applying it to generating tasks where only the history context is given, there is no way to explicitly take into account the future information. Therefore, reranking-based models do not apply here. \cite{li2016deep} addresses future information by policy learning, but the model suffers from high variance due to the enormous sequential search space. \citet{serban2016hierarchical,zhao2017learning,shen2017conditional} adopt the variational inference strategy to reduce the training variance by optimizing over latent continuous variables. However, they all stick to the original MLE objective and no connection with the surrounding context is considered.

In this work, we address both challenges by introducing an auxiliary continuous code space which is learned from the whole dialogue flow. At each time step, instead of directly optimizing discrete utterances, the current, past and future utterances are all trained to maximize the mutual information with this code space. Furthermore, a learnable prior distribution is simultaneously optimized to predict the corresponding code space, enabling efficient sampling in the testing phase without getting access to the ground-truth future conversation. Extensive experiments have been conducted to validate the superiority of our framework. The generated responses clearly demonstrate better performance with respect to both coherence and diversity.
\section{Model Structure}
\subsection{Motivation}
Let $u_i$ be the $i$th utterance within a dialogue flow. The dialogue history $H_{i-1}$ contains all the preceding context $u_1, u_2,\dots,u_{i-1}$ and $F_{i+1}$ denotes the future conversations $u_{i+1},\dots,u_T$. The objective of our model is to find the decoding probability $p_{\theta}(u_i|H_{i-1},F_{i+1})$ that maximizes the mutual information $I(H_{i-1},u_{i})$ and $I(u_i,F_{i+1})$. Formally, the objective is:

\begin{equation}
\label{eq: mmi}
\begin{gathered}
\max_\theta \lambda_{1}I(H_{i-1},u_{i})+\lambda_{2}I(u_i,F_{i+1})\\
u_i\sim p_{\theta}(u_{i}|H_{i-1},F_{i+1})
\end{gathered}
\end{equation}
$\lambda_{1}$ and $\lambda_{2}$ adjusts the relative weight. Mutual information is defined over $p_{\theta}(u_i|H_{i-1},F_{i+1})$ and the empirical distribution $p(H_{i-1},F_{i+1})$. Now we assume the future context $F_{i+1}$ is known to us when training the decoding probability, we will address the unknown future problem later.

Directly optimizing with this objective is unfortunately infeasible because the exact computation of mutual information is intractable, and backpropagating through sampled discrete sequences is notoriously difficult to train. The discontinuity prevents the direct application of the reparameterization trick~\cite{kingma2013auto}. Low-variance relaxations like Gumbel-Softmax~\cite{jang2016categorical}, semantic hashing~\cite{kaiser2018fast} or vector quantization~\cite{van2017neural} lead to biased gradient estimations, which are accumulated as the sequence becomes longer. The Monte-Carlo-Simulation is unbiased but suffers from high variances. Designing a reasonable control variate for variance reduction is an extremely tricky task~\cite{mnih2014neural, tucker2017rebar}. For this sake, we propose replacing $u_i$ with a continuous code space $c$ learned from the whole dialogue flow.

\begin{figure*}[!ht]
\centering
\centerline{\includegraphics[width=14cm,height=8.45cm]{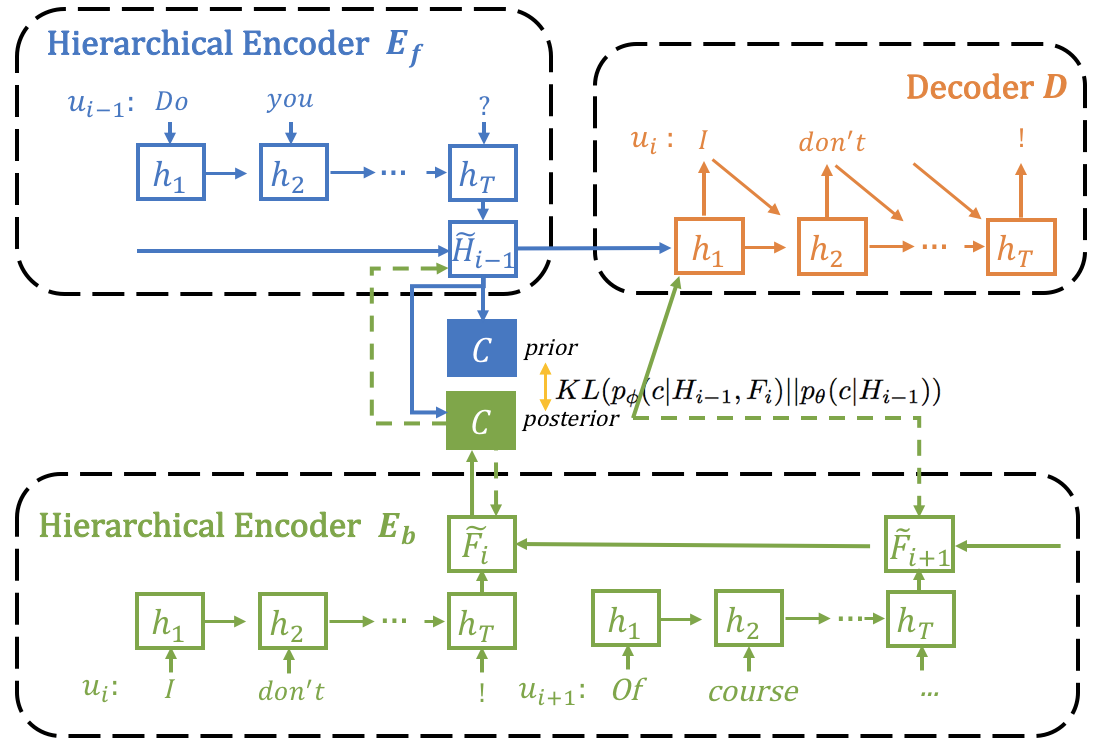}}
\caption{Framework of NEXUS Networks. Full line indicates the generative model to generate the continuous code and corresponding responses. Dashed line indicates the inference model where the posterior code is trained to infer the history, current and future utterances. Both parts are simultaneously trained by gradient descent.}
\label{fig:model}
\end{figure*}
\subsection{Continuous Code Space}
We define the continuous code space $c$ to follow the Gaussian probability distribution with a diagonal covariance matrix conditioning on the whole dialogue:
\begin{equation}
\label{eq: codesample}
c\sim p_{\phi}(c|H_{i-1},F_{i})=\mathcal{N}(\mu,\sigma^2 \mathbb{I}|H_{i-1},F_{i})
\end{equation}
The dialogue history $H_{i-1}$ is encoded into vector $\tilde{H_{i-1}}$ by a forward hierarchical GRU model $E_f$ as in \cite{serban2015building}. The future conversation, including the current utterance, is encoded into $\tilde{F_i}$ by a backward hierarchical GRU $E_b$. $\tilde{H_{i-1}}$ and $\tilde{F_i}$ are concatenated and a multi-layer perceptron is built on top of them to estimate the Gaussian mean and covariance parameters. The code space is trained to infer the encoded history $\tilde{H_{i-1}}$ and future $\tilde{F_{i+1}}$. The full optimizing objective is:
\begin{equation}
\begin{gathered}	
\label{eq: code}
\mathcal{L}(c)=\max_{\phi} \mathbb{E}_{p_{\phi}(H_{i-1},F_{i},c)}[\lambda_{1}\log p_{\phi}(\tilde{H_{i-1}}|c)\\+\lambda_{2}\log p_{\phi}(\tilde{F_{i+1}}|c)]\\
p_{\phi}(H_{i-1},F_{i},c)=p(H_{i-1},F_{i})p_{\phi}(c|H_{i-1},F_{i})\\
p_{\phi}(\tilde{H_{i-1}}|c)=\mathcal{N}(\mu, \sigma^2 \mathbb{I}|c)\\
p_{\phi}(\tilde{F_{i+1}}|c)=\mathcal{N}(\mu, \sigma^2 \mathbb{I}|c)
\end{gathered}
\end{equation}

where $\tilde{H_{i-1}}$ and $\tilde{F_{i+1}}$ are also assumed to be Gaussian distributed given $c$ with mean and covariance estimated from multi-layer perceptrons. We infer the encoded vectors instead of the original sequences for three reasons. Firstly, inferring dense vectors is parallelizable and computationally much cheaper than autoregressive decoding, especially when the context sequences could be unlimitedly long. Secondly, sequence vectors can capture more holistic semantic-level similarity than individual tokens. Lastly, It can also help alleviate the posterior collapsing issue~\cite{bowman2016generating} when training variational inference models on text~\cite{chen2016variational,shen2018improving}, which we will use later. It can be shown that the above objective maximizes a lower bound of $\lambda_{1}I(H_{i-1}, c)+\lambda_{2}I(c,F_{i+1})$, given the conditional probability $p_{\phi}(c|H_{i-1},F_{i})$. The proof is a direct extension of the derivation in \cite{chen2016infogan}, followed by the Data Processing Inequality~\cite{beaudry2012intuitive} that the encoding function can only reduce the mutual information. As the sampling process contains only Gaussian continuous variables, the above objective can be trained through the reparameterization trick~\cite{kingma2013auto}, which is a low-variance, unbiased gradient estimator~\cite{burda2015importance}. After training, samples from $p_{\phi}(c|H_{i-1},F_{i})$ hold high mutual information with both the history and future context. The next step is then transferring the continuous code space to reasonable discrete natural language utterances.
\subsection{Decoding from Continuous Space}
Our decoder transfers the code space $c$ into the ground-truth utterance $u_i$ by defining the probability distribution $p(u_i|H_{i-1},c)$, which is implemented as a GRU decoder going through $u_i$ word by word to estimate the output probability. The encoded history $\tilde{H_{i-1}}$ and code space $c$ are concatenated as an extra input at each time step. The loss function for the decoder is then:
\begin{equation}
\begin{gathered}
\label{eq: decodeobj}
\mathcal{L}(d)=\max_{\phi}\mathbb{E}_{p_{\phi}(H_{i-1},F_{i},c)}\log p_{\phi}(u_i|H_{i-1},c)\\
p_{\phi}(H_{i-1},F_{i},c)=p(H_{i-1},F_{i})p_{\phi}(c|H_{i-1},F_{i})
\end{gathered}
\end{equation}

which can be proved to be the lower bound of the conditional mutual information $I(u_i,c|H_{i-1})$. By maximizing the conditional mutual information, $c_i$ is trained to maintain as much information about the target sequence $u_i$ as possible. 

Combining Eq.~\ref{eq: code} and \ref{eq: decodeobj}, our model until now can be viewed as optimizing a lower bound of the following objective:
\begin{equation}
\label{eq: new-mmi}
\begin{gathered}
\max_{\phi} \lambda_{1}I(H_{i-1}, c)+\lambda_{2}I(c, F_{i+1}) + I(u_i, c|H_{i-1})\\
c\sim p_{\phi}(c|H_{i-1},F_{i})
\end{gathered}
\end{equation}

Compared with the original motivation in Eq.~\ref{eq: mmi}, we sidestep the non-differentiability problem by replacing $u_i$ with a continuous code space $c$, then forcing $u_i$ to contain the same information as maintained in $c$ by additionally maximizing the mutual information between them. 

Nonetheless, Eq.~\ref{eq: new-mmi} and Eq.~\ref{eq: mmi} might lead to different optimums as mutual information does not satisfy the transitive law. In the extreme case, different dimensions of $c$ could individually maintain information about history, current and future conversations and the conversations themselves  do not share any dependency relation. To avoid this issue, we restrict the dimension of $c$ to be smaller than that of the encoded vectors. In this case, optimizing Eq.~\ref{eq: new-mmi} will favor utterances having stronger correlations with the surrounding context to achieve a higher total mutual information.
\subsection{Learnable Prior Distribution for Unknown Future}
The last problem is the sampling mechanism of $c$ in Eq.~\ref{eq: codesample}, which conditions on the ground-truth future conversation. In the testing phase, when we have no access to it, we cannot perform the decoding process as in Eq.~\ref{eq: decodeobj}. To allow for decoding with only the history context, we need to learn an appropriate prior distribution $p_{\theta}(c|H_{i-1})$ for $c$. In the ideal case, we would like
\begin{equation}
p_{\theta}(c|H_{i-1})=\sum_{F_{i}}p_{\phi}(c|H_{i-1},F_{i})=p_{\phi}(c|H_{i-1})
\end{equation}
However, $p_{\phi}(c|H_{i-1})$ is intractable as it integrates over all possible future conversations. We apply variational inference on $c$ to maximize the variational lower bound~\cite{jordan1999introduction}:
\begin{equation}
\begin{gathered}
\label{eq: prior}
\mathcal{L}(p)=\max_{\theta,\phi}\mathbb{E}_{p_{\phi}(c|H_{i-1},F_{i})}\log p_\theta(\tilde{F_{i}}|H_{i-1},c)\\-KL(p_{\phi}(c|H_{i-1},F_{i})||p_\theta(c|H_{i-1}))\\
p_{\theta}(\tilde{F_{i}}|H_{i-1},c)\sim \mathcal{N}(\mu, \sigma^2 \mathbb{I}|H_{i-1},c)\\
p_{\theta}(c|H_{i-1})\sim \mathcal{N}(\mu, \sigma^2 \mathbb{I}|H_{i-1}))
\end{gathered}
\end{equation}
It can be reformulated as maximizing:
\begin{equation}
\begin{gathered}
\label{eq: prior-ref}
\mathbb{E}_{p_{\phi}(c|H_{i-1})}KL( p_\phi(\tilde{F_{i}}|H_{i-1},c)||p_\theta(\tilde{F_{i}}|H_{i-1},c))\\-KL(p_{\phi}(c|H_{i-1})||p_\theta(c|H_{i-1}))
\end{gathered}
\end{equation}
We can see it implicitly matches $p_{\phi}(c|H_{i-1})$ to a tractable Gaussian distribution $p_{\theta}(c|H_{i-1})$ by minimizing the KL divergence between them. It also functions as a regularizer to prevent overfitting when learning $p_\phi(c|H_{i-1},F_{i})$. In the testing phase, we can sample $c$ from the learned prior distribution $p_\theta(c|H_{i-1})$, then generate a response based on it.
\subsection{Summary}
To sum up, the total objective function of our model is:
\begin{equation}
\label{eq: obj}
\mathcal{L}=\mathcal{L}(c)+\mathcal{L}(d)+\mathcal{L}(p)
\end{equation}
Weighting can be added to individual loss functions for better performance, but we find it enough to maintain equal weights and avoid extra hyperparameters. All the parameters are simultaneously updated by gradient descent except for the encoders $E_f$ and $E_b$, which only accept gradients from $\mathcal{L}(d)$ since otherwise the model can easily learn to encode no information for a lower reconstruction loss in $\mathcal{L}(c)$ and $\mathcal{L}(p)$. An overview of our training procedure is depicted in Fig.~\ref{fig:model}.

\section{Relationship to Existing Methods}
\paragraph{MMI decoding}
MMI decoder was proposed by \cite{li2015diversity} and further extended in \cite{wang2017steering}. The basic idea is the same as our model by maximizing the mutual information with the dialogue context. However, the MMI principle is applied only at the testing phase rather than the training phase. As a result, it can only be used to evaluate the quality of a generation by estimating its mutual information with the context. To apply it in a generative task, we have to first sample some candidate responses with the seq2seq model, then rerank them by accounting for the MMI score. Our model differs from it in that we directly estimate the decoding probability thus no post-sampling rerank is needed. Moreover, we further include the future context to strengthen the connection role of the current utterances.
\paragraph{Conditional Variational Autoencoder}
The idea of learning an appropriate prior distribution in Eq.~\ref{eq: prior} is essentially a conditional variational autoencoder~\cite{sohn2015learning} where the accumulated posterior distribution is trained to stay close to a prior distribution. It has also been applied in dialogue generation~\cite{serban2016hierarchical,zhao2017learning}. However, all the above methods stick to the MLE objective function and do not optimize with respect to the mutual information. As we will show in the experiment, they fail to learn the correlation between the utterance and its surrounding context. The generation diversity of these models comes more from the sampling randomness of the prior distribution rather than from the correct understanding of context correlation. Moreover, they suffer from the posterior collapsing problem~\cite{bowman2016generating} and require special tricks like KL-annealing, BOW loss or word drop-out~\cite{shen2018improving}. Our model does not have such problems.
\paragraph{Deep Reinforcement Learning Dialogue Generation}
\cite{li2016deep} first considered future success in dialogue generation and applied deep reinforcement learning to encourage more interactive conversations. However, the reward functions are intuitively hand-crafted. The relative weight for each reward needs to be carefully tuned and the training stage is unstable due to the huge search space. In contrast, our model maximizes the mutual information in the continuous space and trains the prior distribution through the reparamaterization trick. As a result, our model can be more easily trained with a lower variance. Throughout our experiment, the training process of NEXUS network is rather stable and much less data-hungry. The MMI objective of our model is theoretically more sound and no manually-defined rules need to be specified.

\section{Experiments}
\FloatBarrier
\begin{table*}[htbp!]
\begin{center}
  \begin{tabular}{lcccccc}
    \toprule
    \multicolumn{1}{l}{\multirow{2}{*}{\textbf{Model}}} & 
    \multicolumn{3}{c}{\textbf{DailyDialog}} & \multicolumn{3}{c}{\textbf{Twitter}} \\ \cmidrule(l){2-7}
    &\textbf{Average} &\textbf{Greedy} &\textbf{Extreme} & \textbf{Average} &\textbf{Greedy} &\textbf{Extreme} \\
    \midrule
    Greedy &0.443 &0.376\phantom{*}&0.328\phantom{*}&0.510\phantom{*} & 0.341&0.356\phantom{*} \\
	Beam&0.437&0.350\phantom{*}&0.369\phantom{*}&0.505\phantom{*} &0.345 &0.352\phantom{*}\\
	MMI&0.457&0.371\phantom{*}&0.371\phantom{*}& 0.518\phantom{*}& 0.353&0.365\phantom{*}\\
    RL& 0.405& 0.329\phantom{*}&0.305\phantom{*}&0.460\phantom{*} &0.349 &0.323\phantom{*}\\
   VHRED&\textbf{0.491}&0.375\phantom{*}&0.313\phantom{*}&0.525\phantom{*}&0.389&0.372\phantom{*}\\
     
     \hline
   \textbf{NEXUS-H}&0.479&0.381*&\textbf{0.385*}&\textbf{0.558*}&0.392&0.373\phantom{*}\\
   \textbf{NEXUS-F}&0.476&0.383*&0.373\phantom{*}&0.549*& 0.393&0.386*\\ 
   \textbf{NEXUS}&0.488&\textbf{0.392*}&0.384*&0.556*&\textbf{0.397*}&\textbf{0.391*}\\
    \bottomrule
  \end{tabular}
\end{center}
\caption{\label{tab: embed}Results of embedding-based metrics. * indicates statistically significant
difference $(p < 0.05)$ from the best baselines. The same mark is used in Table~\ref{tab: bleu}}
\end{table*}
\subsection{Dataset and Training Details}
We run experiments on the DailyDialog~\cite{li2017dailydialog} and Twitter corpus~\cite{ritter2011data}. DailyDialog contains 13118 daily conversations under ten different topics. This dataset is crawled from various websites for English learner to practice English in daily life, which is high-quality, less noisy but relatively smaller. In contrast, the Twitter corpus is significantly larger but contains more noise. We obtain the dataset as used in \citet{serban2016hierarchical} and filter out tweets that have already been deleted, resulting in about 750,000 multi-turn dialogues. The contents have more informal, colloquial expressions which makes the generation task harder. These two datasets are randomly separated into training/validation/test sets with the ratio of 10:1:1. 

In order to keep our model comparable with the state-of-the-art,
we keep most parameter values the same as in \cite{serban2016hierarchical}. We build our vocabulary dictionary based on the most frequent 20,000 words for both corpus and map other words to a UNK token. The dimensionality of the code space $c$ is 100. We use a learning rate of 0.001 for DailyDialog and 0.0002 for Twitter corpus. The batch size is fixed to 128. The word vector dimension is 300 and is initialized with the public Word2Vec~\cite{mikolov2013efficient} embeddings trained on the Google News Corpus. The probability estimators for the Gaussian distributions are implemented as 3-layer perceptrons with the hyperbolic tangent activation function. As mentioned above, when training NEXUS models, we block the gradient from $\mathcal{L}(c)$ and $\mathcal{L}(p)$ with respect to $E_f$ and $E_b$ to encourage more meaningful encodings. The UNK token is prevented from being generated in the test phase. We implemented all the models with the open-sourced Python library Pytorch~\cite{paszke2017automatic} and optimized using the Adam optimizer~\cite{kingma2014adam}.
\subsection{Compared Models}
We conduct extensive experiments to compare our model against several representative baselines. 

\textbf{Seq2Seq}: Following the same implementation as in \cite{vinyals2015neural}, the seq2seq model serves as a baseline. We try both greedy decoding and beam search~\cite{graves2012sequence} with beam size set to 5 when testing.

\textbf{MMI}: We implemented the bidirectional-MMI decoder as in \citet{li2015diversity}, which showed better performance over the anti-LM model. The hyperparameter $\lambda$ is set to 0.5 as suggested. 200 candidates per context are sampled for re-ranking.

\textbf{VHRED}: The VHRED model is essentially a conditional variational autoencoder with hierarchical encoders~\cite{serban2016hierarchical,zhao2017learning}. To alleviate the posterior collapsing problem, we apply the KL-annealing trick and early stop with the step set as 12,000 for the DailyDialog and 75,000 for the Twitter corpus.

\textbf{RL}: Deep reinforcement learning chatbot as in \cite{li2016deep}. We use all the three reward functions mentioned in the paper and keep the relative weights the same as in the original paper. Policy network is initialized with the above-mentioned MMI model.

\textbf{NEXUS-H}: NEXUS network maximizing mutual information only with the history ($\lambda_2=0$).

\textbf{NEXUS-F}: NEXUS network maximizing mutual information only with the future ($\lambda_1=0$).

\textbf{NEXUS}: NEXUS network maximizing mutual information with both the history and future.

NEXUS-H and NEXUS-F are implemented to help us better analyze the effects of different components in our model. The hyperparameters $\lambda_1$ and $\lambda_2$ in NEXUS are set to be 0.5 and 1 respectively as we find history vector is consistently easier to be reconstructed than the future vector (\ref{app: ratio}).

\FloatBarrier
\begin{table*}[htbp!]
\begin{center}
  \begin{tabular}{lcccccc}
    \toprule
    \multicolumn{1}{l}{\multirow{2}{*}{\textbf{Model}}} & 
    \multicolumn{3}{c}{\textbf{DailyDialog}} & \multicolumn{3}{c}{\textbf{Twitter}} \\ \cmidrule(l){2-7}
    &\textbf{BLEU-1} &\textbf{BLEU-2} &\textbf{BLEU-3} & \textbf{BLEU-1} &\textbf{BLEU-2} &\textbf{BLEU-3} \\
    \midrule
    Greedy &0.394\phantom{*} &0.245&0.157\phantom{*}&0.340\phantom{*} & 0.203\phantom{*}&0.116\phantom{*} \\
	Beam&0.386\phantom{*}&0.251&0.163\phantom{*}&0.338\phantom{*} &0.205\phantom{*} &0.112\phantom{*}\\
	MMI&0.407\phantom{*}&0.269&0.172\phantom{*}& 0.347\phantom{*}& 0.208\phantom{*}&0.118\phantom{*}\\
    RL& 0.298\phantom{*}& 0.186&0.075\phantom{*}&0.314\phantom{*} &0.199\phantom{*} &0.103\phantom{*}\\
   VHRED&0.395\phantom{*}&\textbf{0.281}&0.190\phantom{*}&0.355\phantom{*}&0.211\phantom{*}&0.124\phantom{*}\\
     
     \hline
   \textbf{NEXUS-H}&0.418\phantom{*}&0.279&\textbf{0.199*}&\textbf{0.366*}&0.212\phantom{*}&0.126\phantom{*}\\
   \textbf{NEXUS-F}&0.399\phantom{*} &0.260&0.167\phantom{*}&0.359\phantom{*}& 0.213\phantom{*}&0.123\phantom{*}\\
   \textbf{NEXUS}&\textbf{0.424*}&0.276&0.198*&0.363*&\textbf{0.220*}&\textbf{0.131*}\\
    \bottomrule
  \end{tabular}
\end{center}
\caption{\label{tab: bleu}Results of BLEU score. It is computed based on the smooth BLEU algorithm~\cite{lin2004orange}. p-value interval is computed base on the altered bootstrap resampling algorithm~\cite{riezler2005some}}
\end{table*}

\subsection{Metric-based Performance}
\paragraph{Embedding Score} 
We conducted three embedding-based evaluations (average, greedy and extrema)~\cite{liu2016not}, which map responses into vector space and compute the cosine similarity~\cite{rus2012comparison}. The embedding-based metrics can to a large extent capture the semantic-level similarity between generated responses and ground truth. We represent
words using Word2Vec embeddings trained on
the Google News Corpus. We also measure the uncertainty of the score by assuming each data point is independently Gaussian distributed. The standard deviation yields the $95\%$ confidence interval~\cite{barany2007central}. Table~\ref{tab: embed} reports the embedding scores on both datasets. NEXUS network significantly outperforms the best baseline model in most cases. Notably, NEXUS can absorb the advantages from both NEXUS-H and NEXUS-F. The history and future information seem to help the model from different perspectives. Taking into account both of them does not create a conflict and the combination leads to an overall improvement. RL performs rather poorly on this metric, which is understandable as it does not target the ground-truth responses during training~\cite{li2016deep}.
\paragraph{BLEU Score}
BLEU is a popular metric that
measures the geometric mean of the modified n-gram
precision with a length penalty~\cite{papineni2002bleu}. Table~\ref{tab: bleu} reports the BLEU 1-3 scores. Compared with embedding-based metrics, the BLEU score quantifies the word-overlap between generated responses and the ground-truth. One challenge of evaluating dialogue generation by BLEU score is the difficulty of accessing multiple references for the one-to-many alignment relation. Following \citet{sordoni2015neural,zhao2017learning,shen2018improving}, for each context, 10 more candidate references are acquired by using information retrieval methods (see Appendix~\ref{app: ir} for more details). All candidates are then passed to human annotators to filter unsuitable ones, resulting in 6.74 and 5.13 references for DailyDialog and Twitter dataset respectively. The human annotation is costly, so we evaluate it on 1000 sampled test cases for each dataset. As the BLEU score is not the simple mean of individual sentence scores, we compute the $95\%$ significance interval by bootstrap resampling~\cite{koehn2004statistical,riezler2005some}. As can be seen, NEXUS network achieves best or near-best performances with only greedy decoders. NEXUS-H generally outperforms NEXUS-F as the connection with future context is not explicitly addressed by the BLEU score metric. MMI and VHRED bring minor improvements over the seq2seq model. Even when evaluated on multiple references, RL still performs worse than most models.
\FloatBarrier
\begin{table*}[htbp!]
\centering
\begin{tabular}{l|c|c|c|c|c?c|c|c}
\textbf{Model}&AdverSuc  &Neg-PMI&\#Turns&Distinct-1&Distinct-2&Pri&Post&Flu  \\  \hline
Greedy &0.21$\vert$0.13&47.4$\vert$45.8&0.2$\vert$0.6& .019$\vert$.017 &.096$\vert$.072&0.45&0.04&0.92\\
Beam&0.16$\vert$0.12&47.2$\vert$45.3&0.2$\vert$0.7&.026$\vert$.019 & .103$\vert$.086&0.52&0.06&0.90\\
MMI&0.30$\vert$0.19&45.6$\vert$43.2&1.1$\vert$1.6& .042$\vert$.025& .247$\vert$.117&0.56&0.13&0.89\\
RL&0.13$\vert$0.11&45.0$\vert$42.6&2.3$\vert$2.3& .048$\vert$.033& .324$\vert$.287&0.46&0.15&0.69\\
VHRED&0.19$\vert$0.16&46.8$\vert$44.7&1.7$\vert$1.1&.255$\vert$.106&.431$\vert$.311&0.42&0.22&0.92\\
\hline
\textbf{NEXUS-H}&\textbf{0.36}$\vert$\textbf{0.21}&\textbf{44.1}$\vert$41.8&2.0$\vert$1.8&.263$\vert$.108&.454$\vert$.306&0.66&0.20&0.92\\
\textbf{NEXUS-F}&0.22$\vert$0.12&47.1$\vert$45.9&2.6$\vert$2.2&\textbf{.288}$\vert$.117&.466$\vert$.325&0.51&0.31&\textbf{0.94}\\
\textbf{NEXUS}&0.35$\vert$0.18&44.6$\vert$\textbf{41.4}&\textbf{2.8}$\vert$\textbf{2.5}&.282$\vert$\textbf{.119}&\textbf{.470}$\vert$\textbf{.329}&\textbf{0.70}&\textbf{0.33}&0.93\\
\hline
GROUND&0.87$\vert$0.73&40.5$\vert$38.1&4.8$\vert$4.0&.390$\vert$.215&.522$\vert$.495
&0.92&0.67&0.97\\
\end{tabular}
\caption{{\label{tab: other-performance}}Coherence, diversity and human evaluations. Left: DailyDialog results, right: Twitter results}
\end{table*}
\paragraph{Connecting the preceding} We define two metrics to evaluate the model's capability of  ``connecting the preceding context": \textbf{AdverSuc} and \textbf{Neg-PMI}. AdverSuc measures the coherence of generated responses with the provided context by learning an adversarial discriminator~\cite{li2017adversarial} on the same corpus to distinguish coherent responses from randomly sampled ones. We encode the context and response separately with two different LSTM neural networks and output a binary signal indicating coherent or not\footnote{We apply the same architecture as in \citet{DBLP:journals/corr/LuKZSB17}. In our experiment, the discriminator performs reasonably well in the 4 scenarios outlined in \citet{li2017adversarial} and thus can be used as a fair evaluation metric.}. The AdverSuc value is reported as the success rate that the model fools the classifier into believing its false generations ($p(generated=coherent)>0.5$). Neg-PMI measures the negative pointwise mutual information value $-\log p(c|r)/p(c)$ between the generated response $r$ and the dialogue context $c$. $p(c|r)$ is estimated by training a separate backward seq2seq model. As $p(c)$ is a constant, we ignore it and only report the value of $-\log p(c|r)$. A good model should achieve a higher AdverSuc and a lower Neg-PMI. The results are listed in Table \ref{tab: other-performance}. We can see there is still a big gap between ground-truth and synthesized responses. As expected, NEXUS-H leads to the most significant improvement. MMI model also performs remarkably well, but it requires post-reranking thus the sampling process is much slower. VHRED and NEXUS-F do not help much here, sometimes even slightly degrade the performance. We also tried removing the history context when computing the posterior distribution in VHRED, the resulting model has similar performance among all metrics, which suggests VHRED itself cannot actually learn the correlation pattern with the preceding context. Surprisingly, though RL explicitly set the coherence score as a reward function, its performance is far from satisfying. We assume RL requires much more data to learn the appropriate policy than other models and the training process suffers from a higher variance. The result is thus hard to be guaranteed.
\paragraph{Connecting the following}
We measure the model's capability of ``connecting the following context" from two perspectives: number of the simulated turns and diversity of generated responses. We apply all models to generate multiple turns until a generic response is reached. The set of generic responses is manually examined to include all utterances providing only passive dull replies\footnote{We use a simple rule matching method (see Appendix ~\ref{app: dull}). We manually inspect it on a validation subset and find the accuracy is more than 90\%. Similar methods are adopted in \cite{li2016deep}.}. The number of generated turns can reflect the time that a model can maintain an interactive conversation. The results are reflected in the \textbf{\#Turns} column in Table \ref{tab: other-performance}. As in \cite{li2015diversity}, we measure the diversity by the percentage of distinct unigrams (\textbf{Distinct-1}) and bigrams (\textbf{Distinct-2}) in all generated responses. Intuitively a higher score on these three metrics implies a more interactive generation system that can better connect the future context. Again, NEXUS network dominates most fields.  NEXUS-F brings more impact than NEXUS-H as it explicitly encourages more interactive turns. Most seq2seq models fail to provide an informative response in the first turn. The MMI-decoder does not change much, possibly because the sampling space is not large enough, a more diverse sampling mechanism~\cite{vijayakumar2016diverse} might help. NEXUS network can effectively continue the conversation for 2.8 turns for DailyDialog and 2.5 turns for Twitter, which is closest to the ground truth (4.8 and 4.0 turns respectively). It also achieves the best diversity score in both datasets. It is worth mentioning that NEXUS-H also improves over baselines, though not as significantly as NEXUS-F, so NEXUS is not a trade-off but more like an enhanced version from NEXUS-H and NEXUS-F.

In summary, NEXUS network clearly generates higher-quality responses in both coherence and diversity, even in a rather small dataset like DailyDialog. NEXUS-H contributes more to the coherence and NEXUS-F more to the diversity.
\FloatBarrier
\begin{table*}
\centering
\begin{tabular}{p{43mm}|p{88mm}}
  \textbf{Context} & \textbf{Response}\\
  \hline
  { Hi , Alice , what are you going to do this sunday ?  \_\_eou\_\_ } & \textbf{MMI:} \ Oh , that' s great !  \newline \textbf{VHRED :} \  Well ,  it seems that i'm going to be late . \newline \textbf{RL :} \  How long he it be ? \newline \textbf{NEXUS:} \  Nothing really , what ' s up ?  \\ \hline
  {Did you watch the Oscars on Sunday night ? \_\_eou\_\_ } & \textbf{MMI:} \ I have no idea . \newline \textbf{VHRED:} \ That's not too bad .\newline \textbf{RL :} \  It's my mum's birthday on Sunday . \newline \textbf{NEXUS:} \  No , nothing much. Could you help me ?  \\ \hline
   {Let's go for a drink ! \_\_eou\_\_ } & \textbf{MMI:} \ Thanks . \newline \textbf{VHRED:} \ all right . let me see you then . \newline \textbf{RL :} \  Thank you inviting me .\newline \textbf{NEXUS:} \  Sounds good . Where are we going ? \\ \hline 
    {You won't want to miss our webinar tomorrow !   \_\_eou\_\_ } & \textbf{MMI:} \  lol ! \newline \textbf{VHRED:} \ That ' s terrific !  \newline \textbf{RL :} \  Sure, where where ?\newline \textbf{NEXUS:} \  Thanks for your invitation! I'm free ! \\ \hline
    {How I read the article ?   \_\_eou\_\_ } & \textbf{MMI:} \  there!!! \newline \textbf{VHRED:} \ What ' s good with it ?  \newline \textbf{RL :} \ : I don’t know what to do with it !\newline \textbf{NEXUS:} \   Maybe the force is yourself ! \\ \hline
\end{tabular}
\caption{{\label{tab: examples}}Examples of context-response pairs. \_\_eou\_\_ denotes end-of-utterance. First three rows are from DailyDialog and the last two rows are from Twitter
  }
\end{table*}
\subsection{Human Evaluation}
We also employed crowdsourced judges to provide evaluations for a random sample of 500 items in the DailyDialog test dataset. Participants are asked to assign a binary score to each context-response pair from three perspectives: whether the response coincides with its preceding context (Pri), whether the response is interesting enough for people to continue (Post) and whether the response itself is a fluent natural sentence (Flu). Each sample gets one point if judged as yes and zero otherwise. Each pair is judged by three participants and the score supported by most people is adopted. We also evaluated the inter-annotator consistency by Fleiss'k score\cite{fleiss1971measuring} and obtained k scores of 0.452 for Pri, 0.459 for Post (moderate agreement) and 0.621 for Flu (substantial agreement), which implies most context-response pairs reach a consensus on the evaluation task. We compute the average human score for each model. Unlike metric-based scores, the human evaluation is conducted only on the DailyDialog corpus as it contains less noise and can be more fairly evaluated by human judges. Table \ref{tab: other-performance} shows the result in the last three columns. As can be seen, the pri and post human scores are highly correlated with the automatic evaluation metric ``coherence" and ``\#turns", verifying the validity of these two metrics. As for fluency, there is no significant difference among most models. As we also manually examined, fluency is not a major problem and all models produce mostly well-formed sentences. Overall, NEXUS network does produce responses that are more acceptable to human judges.

Table \ref{tab: examples} presents some randomly sampled context-response pairs provided by MMI, VHRED, RL and NEXUS model. We see NEXUS network does generate more interactive outputs than the other three. Though reranked by the bidirectional language model, the MMI decoder still produces quite a few generic responses. VHRED's utterances are more diverse, but it only cares about answering to the immediate query and makes no efforts to bring about further topics. Moreover, it also generates more inappropriate responses than the others. RL provides diverse responses but sometimes not fluent or coherent enough. We do observe that NEXUS sometimes generate over-complex questions which are not very natural, as in the second example. But in most cases, it outperforms the others.
\section{Conclusion}
In this paper, we propose ``NEXUS Network" to enable more interactive human-computer conversations. The main goal of our model is to strengthen the ``nexus" role of the current utterance, connecting both the preceding and the following dialogue context. We compare our model with MMI, reinforcement learning and CVAE-based models. Experiments show that NEXUS network consistently produces higher-quality responses. The model is easier to train, requires no special tricks and demonstrates remarkable generalization capability even in a very small dataset.

Our model can be considered as combining the objective of MMI and CVAE and is compatible with current improving techniques. For example, mutual information can be maximized under a tighter bound using Donsker-Varadhan or f-divergence representation~\cite{donsker1983asymptotic,nowozin2016f,belghazi2018mutual}. Extending the   code space distribution to more than Gaussian by importance weighted autoencoder~\cite{burda2015importance}, inverse autoregressive flow~\cite{kingma2016improved} or VamPrior~\cite{tomczak2018vae} should also help with the performance.

\section*{Acknowledgments}

We thank all anonymous reviewers, Gerhard Weikum, Jie Zhou, Cheng Niu and the dialogue system team of Wechat AI for valuable comments. Xiaoyu Shen is supported by IMPRS-CS
fellowship. This work is partially funded by DFG collaborative
research center SFB 1102 and Research Grants Council of Hong Kong (PolyU 152036/17E, 152040/18E).

\bibliography{emnlp2018}
\bibliographystyle{acl_natbib_nourl}

\appendix
\onecolumn
\section{Supplementary Material}
\subsection{Proof of Eq.~\ref{eq: code}}

\begin{equation*}
\begin{split}
&\lambda_1I(H,c)+\lambda_2I(c,F)\\
 \geq&\lambda_1I(\tilde{H},c)+\lambda_2I(c,\tilde{F})\\
 =&\lambda_1\mathbb{E}_{p_\phi(\tilde{H}c)}\log \frac{p_\phi(\tilde{H}|c)}{p(\tilde{H})}+\lambda_1\mathbb{E}_{p_\phi(c\tilde{F})}\log \frac{p_\phi(\tilde{F}|c)}{p(\tilde{F})}\\
 =&\lambda_1\mathbb{E}_{p_\phi(\tilde{H}c)}\log p_\phi(\tilde{H}|c)+\lambda_1\mathbb{H}(\tilde{H})+\lambda_2\mathbb{E}_{p_\phi(c\tilde{F})}\log p_\phi(\tilde{F}|c)+\lambda_2\mathbb{H}(\tilde{F})\\
 \geq&\lambda_1\mathbb{E}_{p_\phi(\tilde{H}c)}\log p_\phi(\tilde{H}|c)+\lambda_2\mathbb{E}_{p_\phi(c\tilde{F})}\log p_\phi(\tilde{F}|c)\\
 =&\lambda_1\mathbb{E}_{p_\phi(\tilde{H}c)}\log p_\gamma(\tilde{H}|c)+\lambda_1KL(p_\phi(\tilde{H}|c)||p_\gamma(\tilde{H}|c))+\lambda_2\mathbb{E}_{p_\phi(c\tilde{F})}\log p_\gamma(\tilde{F}|c)+\lambda_2KL(p_\phi(\tilde{H}|c)||p_\gamma(\tilde{H}|c))\\
 \geq&\lambda_1\mathbb{E}_{p_\phi(\tilde{H}c)}\log p_\gamma(\tilde{H}|c)+\lambda_2\mathbb{E}_{p_\phi(c\tilde{F})}\log p_\gamma(\tilde{F}|c)\\
 =&\mathbb{E}_{p_\phi(\tilde{H}u_{i}\tilde{F,c})}[\lambda_1\log p_\gamma(\tilde{H}|c)+\lambda_2\log p_\gamma(\tilde{F}|c)]
 \end{split}
 \end{equation*}
 \subsection{Proof of Eq.~\ref{eq: decodeobj}}
 \begin{equation*}
 \begin{split}
 I(u_i,c|H)&=\mathbb{E}_{p(H)}\mathbb{E}_{p_\phi(u_{i}c|H)}\log\frac{p_\phi(u_i|Hc)}{p(u_i|H)}\\
 &=\mathbb{E}_{p(H)}\mathbb{E}_{p_\phi(u_{i}c|H)}\log p_\phi(u_i|Hc)+\mathbb{H}(u_i|H)\\
 &\geq \mathbb{E}_{p(Hu_{i}F)}\mathbb{E}_{p_\phi(c|Hu_{i}F)}\log p_\phi(u_i|Hc)\\
 &=\mathbb{E}_{p(Hu_{i}F)}\mathbb{E}_{p_\phi(c|Hu_{i}F)}\log p_\gamma(u_i|Hc)+\mathbb{E}_{p_\phi(HcF)}KL(p_\phi(u_i|Hc)||p_\gamma(u_i|Hc))\\
 &\geq \mathbb{E}_{p(Hu_{i}F)}\mathbb{E}_{p_\phi(c|Hu_{i}F)}\log p_\gamma(u_i|Hc)
 \end{split}
 \end{equation*}
 \subsection{Derivation of Eq.~\ref{eq: prior-ref}}
 \begin{equation*}
 \begin{split}
 &\mathbb{E}_{p(\tilde{u_iF}|\tilde{H})}[\mathbb{E}_{p_{\phi}(c|\tilde{H}\tilde{u_{i}F})}\log p_\phi(\tilde{u_{i}F}|c)-KL(p_{\phi}(c|\tilde{H}\tilde{u_{i}F})||p_\theta(c|\tilde{H}))]\\
 =&\mathbb{E}_{p(\tilde{u_iF}|\tilde{H})}[\mathbb{E}_{p_{\phi}(c|\tilde{H}\tilde{u_{i}F})}\log \frac{p_\phi(\tilde{u_iF}|c)p_\theta(c|\tilde{H})}{p_\phi(c|\tilde{H}\tilde{u_iF})}]\\
 =&\mathbb{E}_{p(\tilde{u_iF}|\tilde{H})}[[\mathbb{E}_{p_{\phi}(c|\tilde{H}\tilde{u_{i}F})}\log \frac{p_\phi(\tilde{u_iF}|c)p_\theta(c|\tilde{H})p(\tilde{u_iF}|\tilde{H})}{p_\phi(\tilde{u_iF}|\tilde{H}c)p_\phi(c|\tilde{H})}]\\
 =&\mathbb{E}_{q_\phi(c|\tilde{H})}KL( p_\phi(\tilde{u_{i}F}|\tilde{H}c)||p_\phi(\tilde{u_{i}F}|\tilde{H}c))-KL(p_{\phi}(c|\tilde{H})||p_\theta(c|\tilde{H}))-\mathbb{H}(\tilde{u_iF}|\tilde{H})
 \end{split}
 \end{equation*}

\FloatBarrier
\subsection{Information Retrieval Technique for Multiple References}
\label{app: ir}
We collected multiple reference responses for each dialogue context in the test set by information retrieval techniques. References are retrieved based on their similarity with the provided context. Responses to the retrieved utterances are used as references. The process of retrieving similar context is as follows: First, we select 1000 candidate utterances using the tf-idf score. These candidates are then mapped to a vector space by summing their contained word vectors. After that, they are reranked based on the average of cosine similarity, Jaccard distance and Euclidean distance with the ground-truth context. The top 10 retrieved responses are passed to human annotators to judge the appropriateness.

\subsection{Phrases that count as forming dull responses}
\label{app: dull}
\begin{enumerate}[1)]
\indentitem
\item i know
\item no \_\_eou\_\_(yes \_\_eou\_\_)
\item no problem
\item lol
\item thanks \_\_eou\_\_
\item don't know 
\item don't think 
\item what ?
\item of course
\item wtf
\end{enumerate}
Utterances matching one of these phrases are treated as dull responses.
\subsection{Effect of hyperparameter $\lambda_1 / \lambda_2$}
\label{app: ratio}
\begin{figure*}[!ht]
\centering
\centerline{\includegraphics[width=16cm]{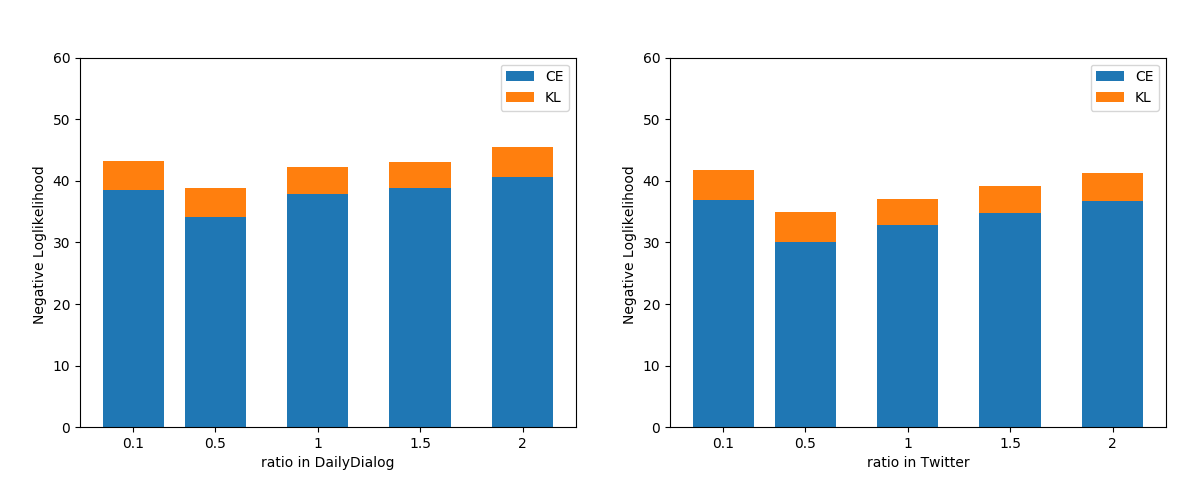}}
\caption{Effect of hyperparameter ratio $\lambda_1 / \lambda_2$ on two datasets. }
\label{fig:parameter}
\end{figure*}
Figure~\ref{fig:parameter} visualizes the effects of hyperparameters $\lambda_1$ and $\lambda_2$. The negative-log-likelihood is decomposed into two parts: decoding cross entropy (CE) as in Eq.~\ref{eq: decodeobj} and KL divergence as in Eq.~\ref{eq: prior}. The sum is a lower bound of the true log-likelihood. The optimal ratio is around 0.5 for both datasets, which means only half weights should be given to the history compared with the future context. Two reasons can explain this phenomena. Firstly, future vector is harder to infer than history as it is not explicitly exposed as an input in Eq.~\ref{eq: code}. Secondly, minimizing the KL divergence in Eq.~\ref{eq: prior} pushes the code space to discard information from the future context so that it could vanish to zero. Therefore, more weights should be given to the future context to maintain a balance.
\end{document}